\documentclass[conference]{IEEEtran}
\IEEEoverridecommandlockouts

\usepackage[T1]{fontenc}
\usepackage{cite}
\usepackage{amsmath,amssymb,amsfonts}
\usepackage{algorithmic}
\usepackage{graphicx}
\usepackage{textcomp}
\usepackage{xcolor}
\def\BibTeX{{\rm B\kern-.05em{\sc i\kern-.025em b}\kern-.08em
    T\kern-.1667em\lower.7ex\hbox{E}\kern-.125emX}}

\usepackage{booktabs}
\usepackage{xurl}
\usepackage{makecell}
\usepackage{multirow}
\usepackage{bm}
\usepackage[hidelinks]{hyperref}
\usepackage[subpreambles=false,mode=buildnew]{standalone}

\newif\ifshellescapeon
\ifnum\pdfshellescape=1
  \shellescapeontrue
\else
  \shellescapeonfalse
\fi

\usepackage{tikz}
\usetikzlibrary{
    arrows.meta,
    calc,
    fit,
    backgrounds
}
\pgfdeclarelayer{background}
\pgfdeclarelayer{foreground}
\pgfsetlayers{background,main,foreground}
\usepackage{listings}
\usepackage{tcolorbox}
\newtcbox{\prompt}{nobeforeafter, boxrule=0.4pt, colback=gray!10, 
                   colframe=gray!40, fontupper=\ttfamily\small,
                   left=2pt, right=2pt, top=1pt, bottom=1pt}

\newcommand\copyrighttext{%
  \footnotesize \textcopyright 2026 IEEE. Personal use of this material is permitted.
  Permission from IEEE must be obtained for all other uses, in any current or future
  media, including reprinting/republishing this material for advertising or promotional
  purposes, creating new collective works, for resale or redistribution to servers or
  lists, or reuse of any copyrighted component of this work in other works.
}
\newcommand\copyrightnotice{%
\begin{tikzpicture}[remember picture,overlay]
\node[anchor=south,yshift=10pt] at (current page.south) {\fbox{\parbox{\dimexpr\textwidth-\fboxsep-\fboxrule\relax}{\copyrighttext}}};
\end{tikzpicture}%
}

\newcommand{\mtg}{\textit{Magic: the Gathering}}

\newcommand{\modelname}[0]{DraftEncoder}
\DeclareMathOperator{\MLP}{MLP}
\DeclareMathOperator{\DO}{DO}
\DeclareMathOperator{\Rope}{R}
\DeclareMathOperator{\Attn}{Attn}
\DeclareMathOperator{\LN}{LN}
\DeclareMathOperator{\LeakyReLU}{LeakyReLU}

\begin{document}
\bstctlcite{IEEEexample:BSTcontrol}

\title{Predicting Drafted Deck Strength for \\ ``Magic: the Gathering''\\
\thanks{This work was supported by JST BOOST, Grant Number JPMJBS2407 and JST CREST, Grant
Number JPMJCR21D1.}
}

\author{\IEEEauthorblockN{Tomas Rigaux}
\IEEEauthorblockA{\textit{Grad. School of Informatics} \\
\textit{Kyoto University}\\
Kyoto, Japan \\
tomas@rigaux.com}
\and
\IEEEauthorblockN{Hisashi Kashima}
\IEEEauthorblockA{\textit{Grad. School of Informatics} \\
\textit{Kyoto University}\\
Kyoto, Japan \\
kashima@i.kyoto-u.ac.jp}}

\maketitle
\copyrightnotice

\IEEEpubidadjcol

\begin{abstract}
Many real-world games do not admit a fixed, compact rule set: instead, their
dynamics are defined by interactions among a large and often evolving collection
of game pieces, making general-purpose policy learning impractical.
\textit{Magic: the Gathering} (MTG) exemplifies this setting, where the cards
themselves define and alter gameplay rules, strategic constraints, and long-term
outcomes, while the pool of available cards is ever-changing.
We study \textit{Draft}, a constrained deck-building format of MTG in which
eight players make 39–45 sequential selections from semi-random packs to
construct a 40-card deck under partial information. By isolating the card
selection process from gameplay, Draft provides a tractable yet non-trivial
setting for studying decision-making driven by combinatorial card synergies. We
propose an encoder-based model that produces set-contextualized card embeddings
to encode the draft decision sequence, with a consistent improvement over linear
baselines on large-scale real-world data, establishing a first learned benchmark
for outcome prediction in MTG Draft. Our code is available at
\href{https://github.com/akulen/MtGDraftEncoder}{github.com/akulen/MtGDraftEncoder}.
\end{abstract}

\begin{IEEEkeywords}
Machine Learning, Card Games, Magic: the Gathering, Neural Networks, Game AI, Transformers.
\end{IEEEkeywords}

\section{Introduction}

Complex games have long served as useful benchmarks for evaluating the
capabilities of learning agents, as they offer clearly defined rules and
measurable objectives. Collectible card games extend the scope as they allow
for the rule set to be altered. The game pieces are a growing collection of
cards, each with the ability to introduce novel mechanics, modify existing
rules, or combine with previous cards to re-contextualize their utility. As a
result, learning a single, global policy that fully captures the game is
impractical: mastery requires continual adaptation to new content and the
ability to encode generic game pieces and how they change the game rules.

In this work, we focus on \textit{Magic: The Gathering} (MTG), a game with
thousands of new cards released every year, and some of the most complex rules
in games. We restrict our attention to the \emph{Draft} format, a variant we
describe in Section~\ref{sec:mtg}, full of highly non-local decisions: the value
of a card depends not only on its intrinsic properties, but also on previously
selected cards, the previously seen cards that could wheel back, and signals
from passed cards giving noisy information about the draft choices made by the
other drafters. We use the outcome of games subsequent to the Draft as an
evaluation metric, making our approach outcome-driven.

Limited work exists on applying machine learning methods to MTG in general,
mostly limited to predicting human pick choices. While such approaches provide
useful signals, they limit themselves to descriptive modeling of player
behavior. In contrast, our goal is to directly predict draft \emph{outcomes}:
given a partial or complete draft sequence, we estimate the expected win-rate
of the resulting deck in subsequent games.

We summarize our contributions as follows:
\begin{enumerate}
    \item A procedure to transform raw card features into contextualized
        embeddings given a draft environment.
    \item \modelname{}, an encoder-based architecture designed to model drafting
        as a sequence of combinatorial decisions, updating card embeddings
        dynamically as the draft progresses, in order to reason about both local
        card quality and global deck composition.
    \item An evaluation of our model and linear baselines on large-scale
        real-world draft data from MTG Arena and
        17Lands~\cite{17lands_17lands_2026}, which we believe is the first
        benchmark for predicting MTG draft performance from draft data.
\end{enumerate}

\section{Magic: the Gathering (MTG)}
\label{sec:mtg}

MTG is the oldest collectible card game, providing a rich history of cards and
ways to play, all of which consist of building a deck following some rules,
then playing with that deck against other players.


\textbf{Drafting process.} During a draft, up to 8 players are arranged around a
table, and they open a pack each, containing a fixed amount of cards (varying
between 13 and 15 depending on the drafted expansion). From their pack
they select exactly one card, pass the remaining cards to the player on their
left, and repeat this process until all cards have been picked. Then this
process is repeated two more times with new packs, alternating the passing
direction from right to left. From the perspective of a single player, the draft
can be viewed as a sequence of $K \in \{39, 42, 45\}$ decisions, where at step
$k$ the player chooses one card from an ever changing candidate set of size
varying between 1 and 15. A choice is informed by previously selected cards, as
well as on implicit signals inferred from which cards remain available in later
picks.

Once all cards have been picked, they are used to construct a deck containing at
least 40 cards, usually including around 17 extra basic lands.


\textbf{Draft representation.} We model a draft for one player as a sequence
$\mathcal{D} = (({(p_{k,i})}_i, c_k))_{k=1}^{K}$, where ${(p_{k,i})}_i$ are the
available cards at pick $k$ and $c_k$ is the selected card.


\begin{table}[t]
    \caption{MTG Expansion Data From 17Lands~\cite{17lands_17lands_2026}, listing the number of distinct cards, the pack size, and number of sample drafts.}
    \label{tbl:17lands}
    \setlength{\tabcolsep}{4pt}
    \begin{minipage}{.5\linewidth}
        \centering
        \begin{tabular}{lccr}
            \toprule
            \bfseries Exp.  & \bfseries Cards & \bfseries Pack & \bfseries Drafts       \\
            \midrule
            \texttt{MOM}         & 346          & 15 & 149,475 \\
            \texttt{WOE}         & 329          & 14 & 125,403 \\
            \texttt{LCI}         & 291          & 15 & 100,720 \\
            \texttt{MKM}         & 326          & 13 & 115,779 \\
            \texttt{OTJ}         & 381          & 14 & 148,496 \\
            \texttt{BLB}         & 276          & 13 & 115,580 \\
            \bottomrule
        \end{tabular}
    \end{minipage}%
    \begin{minipage}{.5\linewidth}
        \centering
        \begin{tabular}{lccr}
            \toprule
            \bfseries Exp.  & \bfseries Cards & \bfseries Pack & \bfseries Drafts       \\
            \midrule
            \texttt{DSK}         & 286          & 14 & 128,212 \\
            \texttt{FDN}         & 286          & 14 & 104,245 \\
            \texttt{DFT}         & 286          & 14 & 101,519 \\
            \texttt{TDM}         & 286          & 14 &  77,374 \\
            \texttt{FIN}         & 363          & 14 & 110,565 \\
            \texttt{EOE}         & 321          & 13 &  79,246 \\
            \bottomrule
        \end{tabular}
    \end{minipage}
\end{table}

\textbf{17Lands draft data.} We use draft data collected from \textit{MTG Arena}
via the 17Lands project~\cite{17lands_17lands_2026}, summarized in
Table~\ref{tbl:17lands}, which aggregates anonymized draft logs from players
across a range of skill levels (The average win rate is typically around 55\%,
inflated by self-selection bias). Each draft record contains the full sequence
of observed packs and picks by an individual player, the player
rank\footnote{Player rank (among \texttt{null}, \texttt{bronze},
\texttt{silver}, \texttt{gold}, \texttt{platinum}, \texttt{diamond}, and
\texttt{mythic}) is a matchmaking tool on \textit{MTG Arena}, to estimate a
combination of player skill and experience in a given draft expansion.}, and the
number of wins and losses.

\textbf{Outcome definition.} We observe the win-rate of the drafted deck as our
measured outcome, stored as the raw number of wins and losses $(W, L)$. For the
data we used, drafted decks are played until either 7 wins or 3 losses are
reached. As a result, the observed proportion of wins is a censored estimate of
the deck’s true win-rate $p_{\text{win}}$.

\textbf{Learning objective.} Given a partial or complete draft sequence
$\mathcal{D}_{\le k}$ in the expansion $\mathcal{E}$, we aim to learn a function
predicting the true win-rate: $f_\mathcal{E}(\mathcal{D}_{\le k}) \rightarrow
p_{\text{win}} \in [0,1]$.

\section{Related Work}

\textbf{Models for card game drafts.} Prior
work~\cite{ward_ai_2021,bertram_learning_2024} on modeling drafting decisions
in \textit{Magic: the Gathering} has primarily focused on estimating pick order
from human draft data. Bertram et al.~\cite{bertram_learning_2024} studies
optimal card representations for draft pick prediction in MTG draft logs.
However, this approach focuses on predicting individual picks rather than
modeling downstream deck performance or win-rate. Some work exists for deriving
a drafting policy combined with a playing policy for
\textit{Hearthstone}~\cite{xiao_mastering_2023}, but relies on being able to
simulate the simpler gameplay.


\textbf{Natural Language Processing.} Converting textual data to a vector of real values
is a recurring task in machine learning. For our work, we use a pre-trained
model, EmbeddingGemma~\cite{vera_embeddinggemma_2025}, that uses a transformer
encoder to encode an input string of unspecified length into a single embedding
of size 768. As drafting is a sequential process where
each decision is highly conditional on both previous decision and previous
potential decisions (which cards were in the passed packs), a model based on
transformer architectures initially introduced by Vaswani
et~al.\cite{vaswani_attention_2017} seems natural.


\section{Our model: \modelname{}}

To predict a draft outcome, we first use a Transformer Encoder architecture to
produce card embeddings contextualized by the current draft expansion and
player rank. Then, the draft state embeddings for each pick are initialized with
those card embeddings. Finally, another attention mechanism updates the draft
state embeddings using the information from previous picks. We give an overview
of this process in Figure~\ref{fig:full_archi}.

\begin{figure}[!ht]
    \centering
  \ifshellescapeon
    \ifdefined\isinputmode
\else
\documentclass{standalone}

\usepackage{tikz}
\usetikzlibrary{
    arrows.meta,
    calc,
    fit,
    backgrounds
}
\pgfdeclarelayer{background}
\pgfdeclarelayer{foreground}
\pgfsetlayers{background,main,foreground}
\usepackage{listings}
\lstset{
   breaklines=true,
   basicstyle=\ttfamily
}

\graphicspath{{figures}} 

\begin{document}
\fi

\begin{tikzpicture}[
    y=-1cm,
    font=\small,
    input/.style={
        align=center,
        minimum width=2.1cm,
        minimum height=1.1cm
    },
    output/.style={
        draw, rounded corners,
        fill=green!25,
        align=center,
        minimum width=2.1cm,
        minimum height=1.1cm
    },
    block/.style={
        draw, rounded corners, thick,
        fill=orange!20,
        align=center,
        minimum width=2.1cm,
        minimum height=1.1cm
    },
    blockI/.style={
        draw, rounded corners=0.4cm, thick,
        fill=blue!10,
        align=center,
        minimum width=1.15cm,
        minimum height=1.1cm
    },
    circ/.style={
        draw, circle, thick,
        fill=white,
        align=center,
        minimum width=0.3cm,
        minimum height=0.3cm
    },
    module/.style={
        draw, thick, rounded corners,
        inner xsep=6pt,
        inner ysep=0pt,
        fill=black!10,
    },
    arrow/.style={->, thick},
]

\node[input] (cardfeat) at (0, 0)
    {Card Features \\ text (+ meta)};

\node[input] (rank) at (3, 0)
    {Rank};

\node[block] (film) at (0, 1.5)
    {FiLM \\ conditioning};

\node[block] (embed) at (3, 1.5) {Embedding};

\draw[arrow] (cardfeat.south) -- (film);
\draw[arrow] (rank.south) -- (embed);
\draw[arrow] (embed) -- (film);

\node[blockI] (rawCard) at (0, 3) {$h_\cdot^{(0)}$};

\draw[arrow] (film) -- (rawCard);

\node (resSplitCA) at (0, 3.878) {};
\node[block] (attnC) at (0, 5.0) {Multi-Head \\ Attention};
\node[circ] (plusC1) at (0, 6.0) {$+$};

\draw[thick] (rawCard) -- (resSplitCA.center);
\draw[arrow, rounded corners] (resSplitCA.center) -| coordinate (cardencLeft) ++(-1.2, 1) |- (plusC1.west);
\draw[arrow] (resSplitCA.center) |- ++(0, 0.3) [rounded corners] -| ($(attnC.north)+(0.7, 0)$);
\draw[arrow] (resSplitCA.center) -- (attnC.north);
\draw[arrow] (resSplitCA.center) |- ++(0, 0.3) [rounded corners] -| ($(attnC.north)-(0.7, 0)$);
\draw[thick] (attnC.south) -- (plusC1.north);

\node (resSplitCF) at (0, 6.6) {};
\node[block] (ffC) at (0, 7.5)
    {Feed \\ Forward};
\node[circ] (plusC2) at (0, 8.645) {$+$};
\node (leftblockbottom) at (0, 8.978) {};

\draw[thick] (plusC1) -- (resSplitCF.center);
\draw[arrow, rounded corners] (resSplitCF.center) -| ++(-1.2, 1) |- (plusC2.west);
\draw[arrow] (resSplitCF.center) -- (ffC);
\draw[thick] (ffC.south) -- (plusC2.north);

\begin{pgfonlayer}{background}
\node[module, fit=(resSplitCA)(cardencLeft)(attnC)(leftblockbottom)] (cardenc) {};
\end{pgfonlayer}
\node at ($(cardenc.west)+(-0.5, 0)$) {$N_\mathcal{E} \times$};

\node[blockI] (setEnc) at (0, 10) {$h_\cdot^{(N_\mathcal{E})}$};

\draw[arrow] (plusC2) -- (setEnc);

\node[input, minimum height=0.5cm] (packs) at (3, 11.5) {Packs};
\node[input, minimum height=0.5cm] (picks) at (6, 11.5) {Picks};

\node[blockI] (packCard) at (3, 10) {$h_{k,i}^{(0)}$};
\node[blockI] (draftState) at (6, 10) {$s_k^{(0)}$};

\draw[arrow] (packs) -- (packCard);
\draw[thick, rounded corners=0.3cm, white, line width=3pt] (setEnc) |- ($(picks)+(0, -0.5)$) -- (draftState);
\draw[arrow, rounded corners=0.3cm] (setEnc) |- ($(packs)+(0, -0.5)$) -- (packCard);
\draw[arrow, rounded corners=0.3cm] (setEnc) |- ($(picks)+(0, -0.5)$) -- (draftState);
\draw[arrow] (picks) -- (draftState);

\node (rightblocktop) at (3, 3.878) {};
\node[block] (attnH) at (3, 7.5) {Multi-Head \\ Attention};
\node[circ] (plusH1) at (3, 6.5) {$+$};

\draw[arrow, rounded corners] (packCard.north) |- ++(-1.2, -0.15) |- (plusH1.west);
\draw[arrow, rounded corners] (packCard.north) |- ++(-0.3, -0.15) -| ($(attnH.south)+(-0.7, 0)$);
\draw[arrow, rounded corners] (draftState.north) |- ++(-0.3, -0.15) -| ($(attnH.south)+(0.0, 0)$);
\draw[arrow, rounded corners] (draftState.north) |- ++(-0.3, -0.15) -| ($(attnH.south)+(0.7, 0)$);
\draw[thick] (attnH.north) -- (plusH1.south);

\node[block, minimum width=1cm, minimum height=0.5cm, fill=red!20] (rope) at (2.65, 8.5) {RoPE};

\node (resSplitHF) at (3, 6.0) {};
\node[block] (ffH) at (3, 5.2)
    {Feed \\ Forward};
\node[circ] (plusH2) at (3, 4.2) {$+$};

\draw[thick] (plusH1) -- (resSplitHF.center);
\draw[arrow, rounded corners] (resSplitHF.center) -| coordinate (draftencLeft) ++(-1.2, -1) |- (plusH2.west);
\draw[arrow] (resSplitHF.center) -- (ffH);
\draw[thick] (ffH.north) -- (plusH2.south);

\node[block, minimum width=0.5cm, minimum height=0.5cm, fill=red!20] (index) at (5.3, 8.5) {$c_k$};
\node[block] (attnS) at (6, 7.5) {Multi-Head \\ Attention};
\node[circ] (plusS1) at (6, 6.5) {$+$};

\draw[arrow, rounded corners=0.3cm] (picks) |- ++(0.6, -0.5) -| ++(0.6, -2) |- (index);
\draw[thick, rounded corners] (plusH2.east) -| ++(0.9, 0.3) |- ($(index.south)+(0,0.2)$) coordinate (draftencSouth) -- (index);
\draw[arrow] (index) -- ($(attnS.south)+(-0.7, 0)$);
\draw[arrow, rounded corners] (index.west) -| ++(-0.2, -0.3) |- (plusS1.west);
\draw[thick, rounded corners, black!10, line width=3pt] (draftencSouth.center) -| ($(attnS.south)+(0.0, 0)$);
\draw[thick, rounded corners, black!10, line width=3pt] (draftencSouth.center) -| ($(attnS.south)+(0.7, 0)$);
\draw[arrow, rounded corners] ($(draftencSouth.center)+(-0.6,0)$) -| ($(attnS.south)+(0.0, 0)$);
\draw[arrow, rounded corners] (draftencSouth.center) -| ($(attnS.south)+(0.7, 0)$);
\draw[thick] (attnS.north) -- (plusS1.south);

\node (resSplitSF) at (6, 6.0) {};
\node[block] (ffS) at (6, 5.2)
    {Feed \\ Forward};
\node[circ] (plusS2) at (6, 4.2) {$+$};

\draw[thick] (plusS1) -- (resSplitSF.center);
\draw[arrow, rounded corners] (resSplitSF.center) -| coordinate (draftencRight) ++(1.2, -1) |- (plusS2.east);
\draw[arrow] (resSplitSF.center) -- (ffS);
\draw[thick] (ffS.north) -- (plusS2.south);
\node (rightblockbottom) at (3, 8.978) {};

\begin{pgfonlayer}{background}
\node[module, fit=(rope)(draftencLeft)(rightblockbottom)(draftencRight)(rightblocktop)] (draftenc) {};
\end{pgfonlayer}
\node at ($(draftenc.east)+(0.5, 0)$) {$\times N_\mathcal{D}$};

\node[blockI] (packCardO) at (3, 3) {$h_{k,i}^{(N_\mathcal{D})}$};
\node[blockI] (draftStateO) at (6, 3) {$s_k^{(N_\mathcal{D})}$};

\draw[arrow] (plusH2) -- (packCardO);
\draw[arrow] (plusS2) -- (draftStateO);

\node[block] (phead) at (6, 1.5) {Predictor \\ Head};
\node[output] (output) at (6, 0) {Output Win \\ Probabilities};

\draw[arrow] (embed) -- (phead);
\draw[arrow] (draftStateO) -- (phead);
\draw[arrow] (phead) -- (output);
\end{tikzpicture}

\ifdefined\isinputmode \fi%
  \else
    \def\isinputmode{}%
    \resizebox{\columnwidth}{!}{\ifdefined\isinputmode
\else
\documentclass{standalone}

\usepackage{tikz}
\usetikzlibrary{
    arrows.meta,
    calc,
    fit,
    backgrounds
}
\pgfdeclarelayer{background}
\pgfdeclarelayer{foreground}
\pgfsetlayers{background,main,foreground}
\usepackage{listings}
\lstset{
   breaklines=true,
   basicstyle=\ttfamily
}

\graphicspath{{figures}} 

\begin{document}
\fi

\begin{tikzpicture}[
    y=-1cm,
    font=\small,
    input/.style={
        align=center,
        minimum width=2.1cm,
        minimum height=1.1cm
    },
    output/.style={
        draw, rounded corners,
        fill=green!25,
        align=center,
        minimum width=2.1cm,
        minimum height=1.1cm
    },
    block/.style={
        draw, rounded corners, thick,
        fill=orange!20,
        align=center,
        minimum width=2.1cm,
        minimum height=1.1cm
    },
    blockI/.style={
        draw, rounded corners=0.4cm, thick,
        fill=blue!10,
        align=center,
        minimum width=1.15cm,
        minimum height=1.1cm
    },
    circ/.style={
        draw, circle, thick,
        fill=white,
        align=center,
        minimum width=0.3cm,
        minimum height=0.3cm
    },
    module/.style={
        draw, thick, rounded corners,
        inner xsep=6pt,
        inner ysep=0pt,
        fill=black!10,
    },
    arrow/.style={->, thick},
]

\node[input] (cardfeat) at (0, 0)
    {Card Features \\ text (+ meta)};

\node[input] (rank) at (3, 0)
    {Rank};

\node[block] (film) at (0, 1.5)
    {FiLM \\ conditioning};

\node[block] (embed) at (3, 1.5) {Embedding};

\draw[arrow] (cardfeat.south) -- (film);
\draw[arrow] (rank.south) -- (embed);
\draw[arrow] (embed) -- (film);

\node[blockI] (rawCard) at (0, 3) {$h_\cdot^{(0)}$};

\draw[arrow] (film) -- (rawCard);

\node (resSplitCA) at (0, 3.878) {};
\node[block] (attnC) at (0, 5.0) {Multi-Head \\ Attention};
\node[circ] (plusC1) at (0, 6.0) {$+$};

\draw[thick] (rawCard) -- (resSplitCA.center);
\draw[arrow, rounded corners] (resSplitCA.center) -| coordinate (cardencLeft) ++(-1.2, 1) |- (plusC1.west);
\draw[arrow] (resSplitCA.center) |- ++(0, 0.3) [rounded corners] -| ($(attnC.north)+(0.7, 0)$);
\draw[arrow] (resSplitCA.center) -- (attnC.north);
\draw[arrow] (resSplitCA.center) |- ++(0, 0.3) [rounded corners] -| ($(attnC.north)-(0.7, 0)$);
\draw[thick] (attnC.south) -- (plusC1.north);

\node (resSplitCF) at (0, 6.6) {};
\node[block] (ffC) at (0, 7.5)
    {Feed \\ Forward};
\node[circ] (plusC2) at (0, 8.645) {$+$};
\node (leftblockbottom) at (0, 8.978) {};

\draw[thick] (plusC1) -- (resSplitCF.center);
\draw[arrow, rounded corners] (resSplitCF.center) -| ++(-1.2, 1) |- (plusC2.west);
\draw[arrow] (resSplitCF.center) -- (ffC);
\draw[thick] (ffC.south) -- (plusC2.north);

\begin{pgfonlayer}{background}
\node[module, fit=(resSplitCA)(cardencLeft)(attnC)(leftblockbottom)] (cardenc) {};
\end{pgfonlayer}
\node at ($(cardenc.west)+(-0.5, 0)$) {$N_\mathcal{E} \times$};

\node[blockI] (setEnc) at (0, 10) {$h_\cdot^{(N_\mathcal{E})}$};

\draw[arrow] (plusC2) -- (setEnc);

\node[input, minimum height=0.5cm] (packs) at (3, 11.5) {Packs};
\node[input, minimum height=0.5cm] (picks) at (6, 11.5) {Picks};

\node[blockI] (packCard) at (3, 10) {$h_{k,i}^{(0)}$};
\node[blockI] (draftState) at (6, 10) {$s_k^{(0)}$};

\draw[arrow] (packs) -- (packCard);
\draw[thick, rounded corners=0.3cm, white, line width=3pt] (setEnc) |- ($(picks)+(0, -0.5)$) -- (draftState);
\draw[arrow, rounded corners=0.3cm] (setEnc) |- ($(packs)+(0, -0.5)$) -- (packCard);
\draw[arrow, rounded corners=0.3cm] (setEnc) |- ($(picks)+(0, -0.5)$) -- (draftState);
\draw[arrow] (picks) -- (draftState);

\node (rightblocktop) at (3, 3.878) {};
\node[block] (attnH) at (3, 7.5) {Multi-Head \\ Attention};
\node[circ] (plusH1) at (3, 6.5) {$+$};

\draw[arrow, rounded corners] (packCard.north) |- ++(-1.2, -0.15) |- (plusH1.west);
\draw[arrow, rounded corners] (packCard.north) |- ++(-0.3, -0.15) -| ($(attnH.south)+(-0.7, 0)$);
\draw[arrow, rounded corners] (draftState.north) |- ++(-0.3, -0.15) -| ($(attnH.south)+(0.0, 0)$);
\draw[arrow, rounded corners] (draftState.north) |- ++(-0.3, -0.15) -| ($(attnH.south)+(0.7, 0)$);
\draw[thick] (attnH.north) -- (plusH1.south);

\node[block, minimum width=1cm, minimum height=0.5cm, fill=red!20] (rope) at (2.65, 8.5) {RoPE};

\node (resSplitHF) at (3, 6.0) {};
\node[block] (ffH) at (3, 5.2)
    {Feed \\ Forward};
\node[circ] (plusH2) at (3, 4.2) {$+$};

\draw[thick] (plusH1) -- (resSplitHF.center);
\draw[arrow, rounded corners] (resSplitHF.center) -| coordinate (draftencLeft) ++(-1.2, -1) |- (plusH2.west);
\draw[arrow] (resSplitHF.center) -- (ffH);
\draw[thick] (ffH.north) -- (plusH2.south);

\node[block, minimum width=0.5cm, minimum height=0.5cm, fill=red!20] (index) at (5.3, 8.5) {$c_k$};
\node[block] (attnS) at (6, 7.5) {Multi-Head \\ Attention};
\node[circ] (plusS1) at (6, 6.5) {$+$};

\draw[arrow, rounded corners=0.3cm] (picks) |- ++(0.6, -0.5) -| ++(0.6, -2) |- (index);
\draw[thick, rounded corners] (plusH2.east) -| ++(0.9, 0.3) |- ($(index.south)+(0,0.2)$) coordinate (draftencSouth) -- (index);
\draw[arrow] (index) -- ($(attnS.south)+(-0.7, 0)$);
\draw[arrow, rounded corners] (index.west) -| ++(-0.2, -0.3) |- (plusS1.west);
\draw[thick, rounded corners, black!10, line width=3pt] (draftencSouth.center) -| ($(attnS.south)+(0.0, 0)$);
\draw[thick, rounded corners, black!10, line width=3pt] (draftencSouth.center) -| ($(attnS.south)+(0.7, 0)$);
\draw[arrow, rounded corners] ($(draftencSouth.center)+(-0.6,0)$) -| ($(attnS.south)+(0.0, 0)$);
\draw[arrow, rounded corners] (draftencSouth.center) -| ($(attnS.south)+(0.7, 0)$);
\draw[thick] (attnS.north) -- (plusS1.south);

\node (resSplitSF) at (6, 6.0) {};
\node[block] (ffS) at (6, 5.2)
    {Feed \\ Forward};
\node[circ] (plusS2) at (6, 4.2) {$+$};

\draw[thick] (plusS1) -- (resSplitSF.center);
\draw[arrow, rounded corners] (resSplitSF.center) -| coordinate (draftencRight) ++(1.2, -1) |- (plusS2.east);
\draw[arrow] (resSplitSF.center) -- (ffS);
\draw[thick] (ffS.north) -- (plusS2.south);
\node (rightblockbottom) at (3, 8.978) {};

\begin{pgfonlayer}{background}
\node[module, fit=(rope)(draftencLeft)(rightblockbottom)(draftencRight)(rightblocktop)] (draftenc) {};
\end{pgfonlayer}
\node at ($(draftenc.east)+(0.5, 0)$) {$\times N_\mathcal{D}$};

\node[blockI] (packCardO) at (3, 3) {$h_{k,i}^{(N_\mathcal{D})}$};
\node[blockI] (draftStateO) at (6, 3) {$s_k^{(N_\mathcal{D})}$};

\draw[arrow] (plusH2) -- (packCardO);
\draw[arrow] (plusS2) -- (draftStateO);

\node[block] (phead) at (6, 1.5) {Predictor \\ Head};
\node[output] (output) at (6, 0) {Output Win \\ Probabilities};

\draw[arrow] (embed) -- (phead);
\draw[arrow] (draftStateO) -- (phead);
\draw[arrow] (phead) -- (output);
\end{tikzpicture}

\ifdefined\isinputmode \fi}%
  \fi

    \caption{
        Overview of \modelname{}'s full architecture. The main block on the
        left encodes cards in a given expansion setting as described in
        Subsection~\ref{model:enc_set}, and is iterated $N_\mathcal{E}$ times
        to output $h_c^{(N_\mathcal{E})}$ for each card $c$. The main block on
        the right uses these card encodings to encode the draft process as
        described in Subsection~\ref{model:enc_draft}, and is iterated
        $N_\mathcal{D}$ times to output the final draft states
        $s_k^{(N_\mathcal{D})}$ for each pick step $k$. The packs $p_{k,\cdot}$
        are used to select the relevant card states initially, and the picks
        $c_k$ are used at each iteration (as the red box labeled $c_k$) to
        choose the appropriate embedding to update the draft states.
    }
    \label{fig:full_archi}
\end{figure}
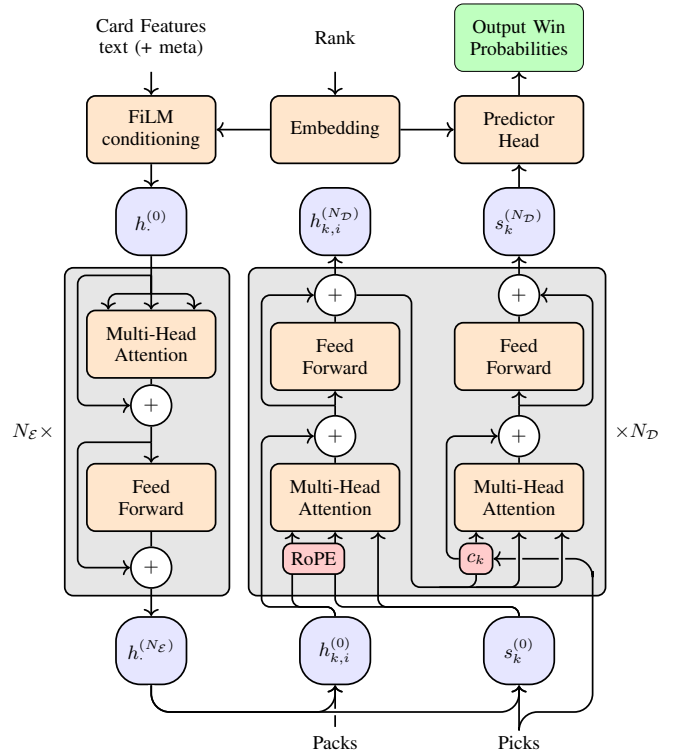

\subsection{Encoding Cards}
\label{model:enc_set}

\begin{figure}[!t]
    \centering
  \ifshellescapeon
    \includestandalone[width=0.9\columnwidth]{figures/card_embedding}%
  \else
    \def\isinputmode{}%
    \resizebox{0.9\columnwidth}{!}{\input{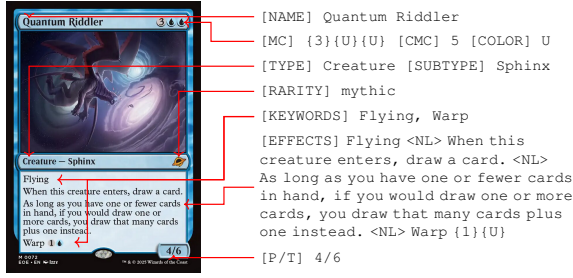}}%
  \fi

    \caption{
        To prepare the oracle string of a card, we concatenate the applying
        information among: Name, Costs, Rarity, Types, Power/Toughness/Loyalty,
        Keywords, and Effects; taken from the updated oracle version found on
        Scryfall~\cite{scryfall_scryfall_2026}. If a card has more than one
        face, we include them both, indicated with the tag \texttt{[FACE\_<i>]}.
    }
    \label{fig:card-emb}
\end{figure}

To construct card features, we convert all the information in a card into a
string (see \figurename~\ref{fig:card-emb}) and use EmbeddingGemma
\cite{vera_embeddinggemma_2025} to convert it to a static embedding vector
$E_c^{(T)}$ of size $d_t=768$. To evaluate the model comprehension, we also
compute statistical features $E_c^{(N)}$ for each card, following the
\texttt{\sc{meta}} features of Bertram et al.~\cite{bertram_learning_2024} (i.e.
the average card win rate when it is in the opening hand, drawn, tutored, in the
deck, in the sideboard, or seen during the game). In our results, we indicate
when a model concatenates these \texttt{\sc{meta}} features to the textual
features.

We reduce the dimension of the initial card features using a
linear layer, a LeakyReLU non-linearity, and a dropout layer.
\begin{align*}
    \bar h_c^{(0)} &= \DO\left(\LeakyReLU\left(W^I \left[E_c^{(T)}||E_c^{(N)}\right] + b^I\right)\right)
\end{align*}

To let player rank condition card evaluation, we learn an embedding vector for
each rank, feed it through a two-layer Multi-Layer Perceptron (MLP) to get two
vectors $\gamma,\beta\in\mathbb{R}^d$. We use each dimension of those vectors to
alter the matching dimension of the card features individually, following the
FiLM layer~(Feature-wise Linear Modulation) of Perez
et~al.~\cite{perez_film_2018}, to control mean and variance, yielding our
initial card embeddings $h_\cdot^{(0)}$:
\begin{align*}
    \gamma^*(r), \beta(r) &= \MLP(x_r) \quad \gamma(r) = 1 + 0.1 \tanh(\gamma^*(r)) \\
    h_c^{(0)} &= \gamma(r) \odot \bar h_c^{(0)} + \beta(r)
\end{align*}

$h_c^{(0)}$ should encode absolute information about card $c$, contextualized by
the skill level of the drafting player, and optionally by empirical meta-stats
about that card in the currently drafted expansion, but is mostly agnostic
relative to other cards in the same expansion. To update this encoding with the
context of the current expansion and other cards that could be drafted with it,
we implement a standard encoder-only architecture in which each card attends to
other cards in the same expansion. This consists of $N_\mathcal{E}$ residual
attention layers, producing contextualized card embeddings
$h_c^{(N_\mathcal{E})}$.

\subsection{Encoding draft picks}
\label{model:enc_draft}

The sequence of information gained during a draft and the decisions taken is
encoded using two types of tokens:
\begin{itemize}
    \item \textbf{Pack cards}: For each pack card $p_{k, i}$, we keep track of
        our evaluation of that card relative to the previous picks, starting
        with the feature vector $h_{k, i}^{(0)}=h_{p_{k, i}}^{(N_\mathcal{E})}$.
    \item \textbf{Draft state}: We also track the current state of the draft
        after $k$ picks, starting by only considering the picked cards in
        isolation with the feature vector
        $s_k^{(0)}=h_{p_{k,c_k}}^{(N_\mathcal{E})}$
\end{itemize}
Those encodings are updated with a bidirectional pre-norm Transformer encoder
block, first updating the pack cards:
\begin{align*}
    \bar h_{k,i}^{(n+1)} &= h_{k,i}^{(n)} + \DO(\Attn(\Rope(\LN(h_{k,i}^{(n)})), \Rope(s_{1\ldots k}^{(n)}), s_{1\ldots k}^{(n)})) \\
    h_{k,i}^{(n+1)} &= \bar h_{k,i}^{(n+1)} + \DO(\MLP(\LN(\bar h_{k,i}^{(n+1)})))
\end{align*}
applying a rotary positional encoding~\cite{su_roformer_2024} $\Rope$ to the query and key,
before updating the draft states:
\begin{align*}
    \bar s_k^{(n+1)} &= s_k^{(n)} + \DO(\Attn(h_{k,c_k}^{(n+1)}, h_{k,\cdot}^{(n+1)}, h_{k,\cdot}^{(n+1)})) \\
    s_k^{(n+1)} &= \bar s_k^{(n+1)} + \DO(\MLP(\LN(\bar s_k^{(n+1)})))
\end{align*}
to obtain $s_k^{(n+1)}$.

After $N_\mathcal{D}$ such updates, we obtain final draft states
$s_k^{(N_\mathcal{D})}$, that we concatenate to the rank embeddings $x_r$ and
feed through an MLP with 3 layers, $\LeakyReLU$ activations, and a final sigmoid
to output after each pick a win-rate between 0 and 1.

\subsection{Training metric}

As our target outcome is biased to have at most 7 wins and 3 losses, we train
our model with an appropriate Negative Log-Likelihood (NLL) loss.
\begin{align*}
    \ell(p_i; W, L)
    &=
    \begin{cases}
    -\log \left[ \binom{6 + L}{6} \, p^{7} (1 - p)^{L} \right] & \text{if } W = 7, \\
    -\log \left[ \binom{W + 2}{2} \, p^{W} (1 - p)^{3} \right] & \text{if } L = 3 ,
    \end{cases} \\
    L(p, W, L) &= \sum_{i=1}^K \frac{i}{K} \ell(p_i; W, L)
\end{align*}
We weight later predictions higher as they were made with more information, and
$p$ is the win-rate predicted by the model.

\section{Experimental Setup}

\subsection{Experiments}

\textbf{Predictive accuracy.} We first evaluate our model on predicting draft
outcomes using the same expansions it was trained on. We only use \texttt{BLB},
\texttt{FDN}, and \texttt{TDM} for this experiment, split temporally using the
last 6 weeks of data into chunks of 2 weeks assigned to training, validation,
and testing chronologically, for around 120,000 training drafts.

\textbf{New expansion and cards generalization.} We also evaluated the ability
of our model to generalize to unseen expansions and cards. To do so, we trained
and validated on the last 4 weeks of every expansion listed in
Table~\ref{tbl:17lands} except \texttt{FDN}, and evaluated on the last 2 weeks
of \texttt{FDN}, for around 400,000 training drafts. Then, we fine-tune each
model using the first four days of data of \texttt{FDN}.

\subsection{Evaluated Models}

We compare \modelname{} against these baselines:
\begin{itemize}
    \item {\sc Heuristic}, that computes and outputs the average "in deck" win
        rate from 17Lands of the best 23 picked cards.
    \item {\sc LR}, a linear regression, using the features of the first $i$
    picked cards and an embedding player rank vector to predict the win rate
    after $i$ picks.
\end{itemize}
We train the linear regression and our model for 100 epochs, with an inner
dimension of 64, $N_\mathcal{E}=10$, $N_\mathcal{D}=5$, 2 attention heads, a
batch size of 256, optimizing with AdamW~\cite{loshchilov_decoupled_2019},
and a peak learning rate of 1e-4.

\section{Results}

We report, as mean $\pm$ standard deviation over 3 random seeds for the epoch
with the best validation loss, the Spearman's rank correlation coefficient
between the predicted win-rates and the observed (biased) win-rates
$\frac{W}{W+L}$\footnote{We note that the label noise inherent to the sampled
data limits the achievable correlation of any model. The correlation between a
true win rate probability $p\sim\mathcal{N}(0.53, 0.05)$ (calibrated to match
the observed distribution via simulation) and an observed win-rate generated
with i.i.d. Bernoulli trials with success probability $p$ under a 7-win/3-loss
stopping rule is around $0.23$}. We use rank correlation since we care about
relative deck strength rather than calibrated probability estimates. Our models
were coded in Jax~\cite{bradbury_jax_2018} and
Equinox~\cite{kidger_equinox_2021}, and trained on 4 Nvidia RTX A5000 GPUs,
using around an hour per seed.

\begin{figure}[!t]
    \centering
    \includegraphics[width=\columnwidth]{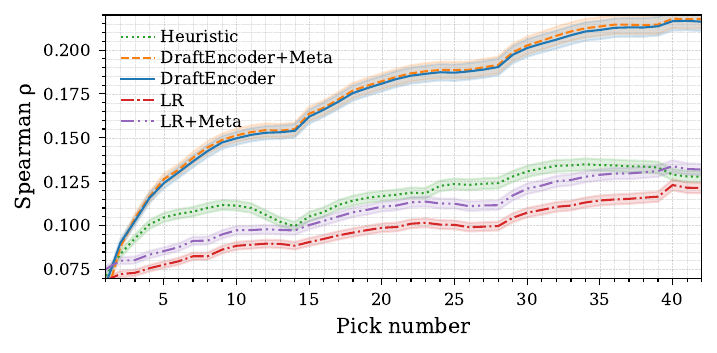}
    \caption{
        Spearman $\rho$ across pick number, on \texttt{BLB}, \texttt{FDN}, and
        \texttt{TDM}. Shaded bands show $\pm$ standard deviation over 3 seeds.
    }
    \label{fig:spear-same}
\end{figure}

As seen in \figurename~\ref{fig:spear-same}, \modelname{} is able to use its
non-linear architecture to outperform the baselines. Although \modelname{} has
slightly worse predictions for the first pick, it quickly grows to almost twice
the rank correlation of the baselines, and is close to reaching the saturation
value of around $0.23$. Adding meta features consistently improves performance,
which we assume means our model is able to learn enough about the cards from the
training outcomes alone.

However, as shown in Table~\ref{tbl:res-new-spear}, the model struggles to
generalize to new sets, without having access to the meta features, only
matching the performance of the linear model without meta features (and falling
short of the heuristic, which has implicit access to meta statistics for the
full expansion), but does manage to be competitive with the linear model with
meta features. Fine-tuning using the first four days of data lowers the
importance of the meta features a lot, as \modelname{} then outperforms an
untuned \modelname{}+Meta, and is only worse than the tuned version by 10\%.
This shows that our model can learn a lot with only a few days of data, making
it useful and relevant early in the lifespan of a newly released expansion.

\begin{table}[t]
    \caption{Spearman coefficient when tested on the last two weeks of
    \texttt{FDN} and trained on all other expansions, and fine-tuned on the
    first four days of \texttt{FDN}.}
    \label{tbl:res-new-spear}
    \centering
    \begin{tabular}{lccc}
        \toprule
        \multirow{2}{*}{\bfseries Model} & \multicolumn{2}{c}{\bfseries Test loss (Spearman $\rho$)} \\
                          & No fine-tuning                  & Four days                       \\
        \midrule
        Heuristic         & $\underline{0.1391 \pm 0.0203}$ & --                              \\
        LR                & $0.0690 \pm 0.0031$             & $0.0694 \pm 0.0041$             \\
        LR+Meta           & $0.0886 \pm 0.0074$             & $0.0898 \pm 0.0081$             \\
        \midrule
        \modelname{}      & $0.0879 \pm 0.0131$             & $\underline{0.1641 \pm 0.0318}$ \\
        \modelname{}+Meta & $\bm{0.1529 \pm 0.0262}$        & $\bm{0.1809 \pm 0.0381}$        \\
        \bottomrule
    \end{tabular}
\end{table}

\section{Conclusion}

In this work, we introduced \modelname{}, the first learned benchmark for
predicting \mtg{} draft outcomes. Our model outperforms baselines by nearly
2$\times$ in rank correlation, and approaches the estimated noise ceiling of
$0.23$ imposed by sampling noise inherent to the collected data.

The main remaining challenge to address is generalization to unseen expansions
and cards, though fine-tuning on as little as four days of initial data largely
closes the gap to meta-augmented models, suggesting that our model learns
transferable draft representations that adapt quickly to new cards. Future work
could explore incorporating gameplay simulation or a stronger natural language
processing (NLP) of the cards, potentially including a fine-tuned rather than
frozen text encoder.

\bibliography{biblio_cfg,biblio}
\bibliographystyle{IEEEtran}
\end{document}